\documentclass[nohyperref]{article}

\usepackage{microtype}
\usepackage{graphicx}
\usepackage{subfigure}
\usepackage{booktabs} 
\usepackage{hyperref}

 \usepackage[accepted]{icml2022}

\usepackage{amsmath}
\usepackage{amssymb}
\usepackage{mathtools}
\usepackage{amsthm}

\usepackage{rotating}
\usepackage{multirow}
\usepackage{bbm}
\usepackage{hyperref}
\usepackage{url}
\usepackage{float}

\usepackage[capitalize,noabbrev]{cleveref}

\theoremstyle{plain}

\theoremstyle{definition}

\theoremstyle{remark}

\usepackage[textsize=tiny]{todonotes}

\icmltitlerunning{Learning Graph Augmentations to Learn Graph Representations}

\begin{document}

\twocolumn[
\icmltitle{Learning Graph Augmentations to Learn Graph Representations}



\icmlsetsymbol{equal}{*}

\begin{icmlauthorlist}
\icmlauthor{Kaveh Hassani}{yyy}
\icmlauthor{Amir Hosein  Khasahmadi}{yyy}
\end{icmlauthorlist}

\icmlaffiliation{yyy}{Autodesk AI Lab, Toronto, Canada}
\icmlcorrespondingauthor{Kaveh Hassani}{kaveh.hassani@autodesk.com}

\icmlkeywords{Machine Learning, ICML}

\vskip 0.3in
]



\printAffiliationsAndNotice{}  

\begin{abstract}
Devising augmentations for graph contrastive learning is challenging due to their irregular structure, drastic 
distribution shifts, and nonequivalent feature spaces across datasets. We introduce LG2AR, \textbf{L}earning
\textbf{G}raph \textbf{A}ugmentations to \textbf{L}earn \textbf{G}raph \textbf{R}epresentations, which is an 
end-to-end automatic graph augmentation framework that helps encoders learn generalizable representations on 
both node and graph levels. LG2AR consists of a probabilistic policy that learns a distribution over 
augmentations and a set of probabilistic augmentation heads that learn distributions over augmentation 
parameters. We show that LG2AR achieves state-of-the-art results on 18 out of 20  graph-level and node-level 
benchmarks compared to previous unsupervised models under both linear and semi-supervised evaluation protocols.
\end{abstract}

\section{Introduction}
Graph Neural Networks (GNNs) are a class of deep models that learn node representations over order-invariant 
and variable-size data, structured as graphs, through an iterative process of transferring, transforming, and 
aggregating the representations from topological neighbors. The learned representations are then summarized 
into a graph-level representation 
\citep{li_2015_iclr, gilmer_2017_icml, kipf_2017_iclr, velickovic_2018_iclr, xu_2019_iclr, Khasahmadi2020Memory}. 
GNNs are applied to non-Euclidean data such as point clouds \citep{ hassani_2019_iccv}, robot designs 
\citep{wang_2018_iclr}, physical processes \citep{pmlr-v119-sanchez-gonzalez20a}, molecules 
\citep{duvenaud_2015_nips}, social networks \citep{kipf_2017_iclr}, and knowledge graphs \citep{vivona_2019_nips}.

GNNs are mostly trained end-to-end with supervision from task-dependent labels. Nevertheless, annotating graphs 
is more challenging compared to other common modalities because they usually represent concepts in specialized 
domains such as biology where labeling through wet-lab experiments is resource-intensive \citep{you2020graph,Hu2020Strategies} 
and labeling them procedurally using domain knowledge is costly\citep{Sun2020InfoGraph}. To address this, unsupervised 
objectives are coupled with GNNs to learn representations without relying on labels which are transferable to a priori 
unknown down-stream tasks. Graph AutoEncoders (GAE) preserve the topological closeness of the nodes in the representations 
by forcing the model to recover the neighbors from the latent space \citep{kipf_2016_arxiv, duran_2017_nips, pan_2018_ijcai, park_2019_iccv}. GAEs over-emphasize proximity 
information at the cost of structural information \citep{velickovic_2019_iclr}. Contrastive methods train graph encoders 
by maximizing the Mutual Information (MI) between node-node, node-graph, or graph-graph representations achieving state-of-the-art results 
on both node and graph classification benchmarks\citep{velickovic_2019_iclr,Sun2020InfoGraph, pmlr-v119-hassani20a, you2020graph, zhu2021graph}. 
For a review on graph contrastive learning see \citep{xie2021self, liu2021graph}.

Contrastive learning is essentially learning invariances to data augmentations which are thoroughly explored for Computer Vision (CV) 
\citep{shorten2019survey} and Natural Language Processing (NLP) \citep{50311}. Learning policies to sample dataset-conditioned augmentations 
is also studied in CV \citep{hataya2020faster, AutoAugment, Cubuk_2019_CVPR, Yonggang2020}. The irregular structure of graphs complicates 
both adopting augmentations used on images and also devising new augmentation strategies \citep{zhao2021data}. Unlike image datasets 
where the distribution is mostly from natural images, graph datasets are abstractions diverse in nature and contain shifts on marginal/conditional 
distributions and nonequivalent feature spaces across datasets \citep{hassani2022crossdomain}. This 
implies that the effect of augmentations is different across the datasets and hence both augmentations and their selection policy should be 
learned from the data to adapt to new datasets.

\textbf{Present Work}. We introduce LG2AR, \textbf{L}earning \textbf{G}raph \textbf{A}ugmentations to \textbf{L}earn \textbf{G}raph \textbf{R}epresentations, a fully-automated 
end-to-end contrastive learning framework that helps encoders learn transferable representations. Specifically, our 
contributions are as follows: (1) We introduce a probabilistic augmentation selection policy that learns a distribution over the augmentation 
space conditioned on the dataset to automate the combinatorial augmentation selection process. (2) We introduce probabilistic augmentation 
heads where each head learns a distribution over the parameters of a specific augmentation to learn better augmentations for each dataset. (3) 
We train the policy and the augmentations end-to-end without requiring an outer-loop optimization and show that unlike other 
methods, our approach can be used for both node-level and graph-level tasks. Finally, (4) we exhaustively evaluate our approach under linear 
and semi-supervised evaluation protocols and show that it achieves state-of-the-art results on 18 out of 20 graph and node level classification 
benchmarks.

\section{Related Work}
Graph augmentations are explored in supervised settings to alleviate over-smoothing and over-fitting. DropEdge \citep{Rong2020DropEdge} 
randomly drops a fraction of the edges during training. ADAEDGE \citep{chen2020measuring} learns to perturb edges between based on the 
predicted node classes. BGCN \citep{zhang2019bayesian} generates an ensemble of deonised graphs by perturbing edges. GAUG \citep {zhao2021data} 
trains a GAE to generate edge probabilities and interpolates them with the original connectivity to sample a denoised graph. FLAG 
\citep{kong2020flag} augments node features with adversarial perturbations and GraphMask \citep{schlichtkrull2021interpreting} introduces 
differentiable edge masking to achieve interpretability. These works assume that a specific type of augmentation suffice for all supervised tasks 
and do not utilize the benefit of mixing the augmentations.

Graph augmentations are also studied in contrastive setting. DGI \citep{velickovic_2019_iclr} 
and InfoGraph \citep{Sun2020InfoGraph} adopt DeepInfoMax \citep{hjelm_2019_iclr} and enforce the consistency between local (node) and 
global (graph) representation. MVGRL \citep{pmlr-v119-hassani20a} augments a graph via graph diffusion and constructs two views by 
randomly sampling sub-graphs from the adjacency and diffusion matrices. GCC \citep{qiu2020gcc} uses sub-graph instance discrimination 
and contrasts sub-graphs from a multi-hop ego network. GraphCL \citep{you2020graph} uses trial-and-error to hand-pick graph augmentations 
and the corresponding parameters of each augmentation. JOAO \citep{you2021graph} extends the GraphCL using a bi-level min-max optimization 
that learns to select the augmentations. Nevertheless, it does not show much improvement over GraphCL. GRACE \citep{zhu2020deep} uses a 
similar approach to GraphCL to learn node representations. GCA \citep{zhu2021graph} uses a set of heuristics to adaptively pick the augmentation 
parameters. BGRL \citep{thakoor2021bootstrapped} adopts BYOL \citep{NEURIPS2020_f3ada80d} and uses random augmentations to learn node 
representations. ADGCL \citep{suresh2021adversarial} introduces adversarial graph augmentation strategies to avoid capturing redundant 
information. Different from prior methods, LG2AR emphasizes the importance of conditional augmentations and learns a distribution over the 
augmentation space along with a set of distributions over the augmentation parameters end-to-end without requiring a bi-level optimization and 
outperforms the previous contrastive methods on both node and graph level benchmarks under linear and semi-supervised evaluation protocols.

\begin{figure*}[ht] 
\begin{center}
\centerline{\includegraphics[width=175mm]{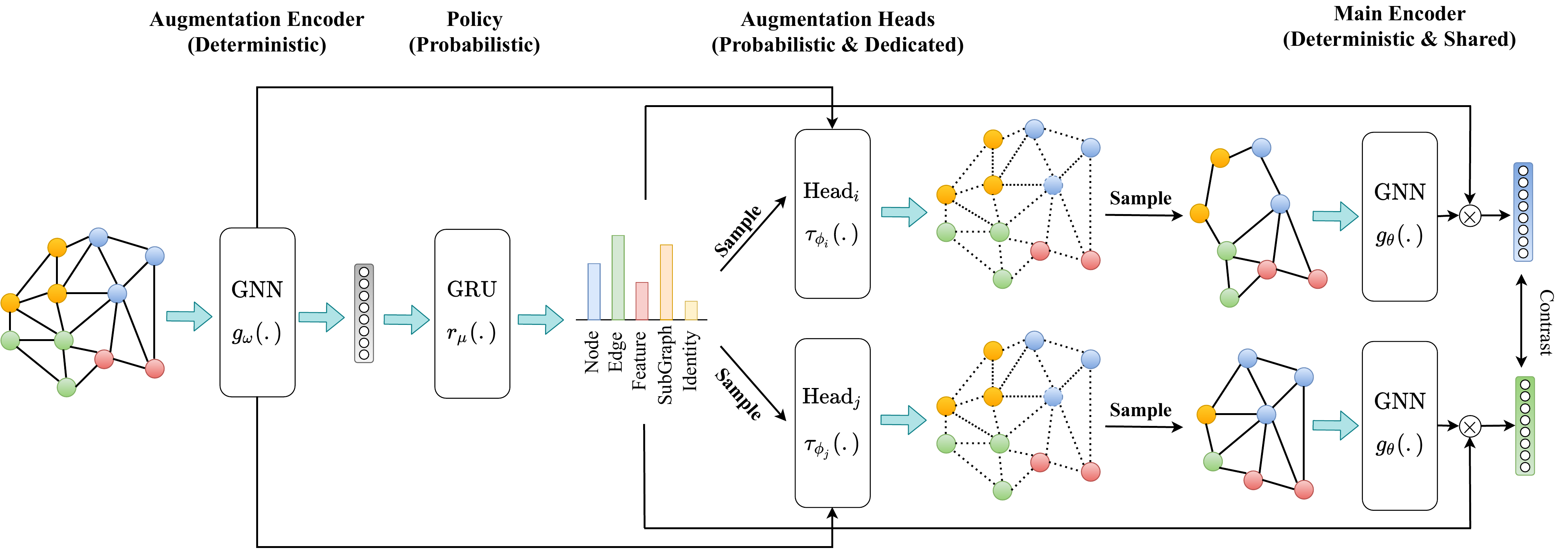}}
\caption{The proposed framework for learning graph augmentation end-to-end.}\label{fig:arch}
\end{center}
\end{figure*}

\section{Method}
Given a dataset of graphs $\mathcal{G}=\left\{G_k\right\}_{k=1}^N$ where each graph $G_k = \left(\mathcal{V}, \mathcal{E}, \textbf{X}\right)$ consists of $|\mathcal{V}|$ nodes, $|\mathcal{E}|$ edges ($\mathcal{E} \subseteq \mathcal{V} \times \mathcal{V}$), and initial node 
features $\textbf{X}\in \mathbb{R}^{|\mathcal{V}| \times d_x}$, and assuming that the semantic labels are not available during the training, the goal is to learn node-level 
representations $\textbf{H}_v\in \mathbb{R}^{|\mathcal{V}| \times d_h}$ and graph-level representation $h_G \in \mathbb{R}^{d_h}$ such that the learned representations are transferable to the 
down-stream tasks unknown a priori. Assuming a set of possible rational augmentations $\mathcal{T}$over $\mathcal{G}$ where each augmentation 
$\tau_{i} \in\mathcal{T}, i\in \{1,...,|\mathcal{T}|\}$ is defined as a function over graph $G_k$ that generates an identity-preserving view of the graph: $G^i_k=\tau_{i}(G_k)$, a  
contrastive framework with negative sampling strategy uses $\mathcal{T}$to draw positive samples from the joint distribution $p\left(\tau_{i}(G_k), \tau_{j}(G_k)\right)$ 
in order to maximize the agreement between different views of the same graph $G_k$ and to draw negative samples from the product of 
marginals $p\left(\tau_{i}(G_k)\right) \times p\left(\tau_{j}(G_{k'})\right)$ to minimize it for views from two distinct graphs $G_k$ and $G_{k'}, k\neq k'$. Most works, sample the 
augmentations uniformly and use trail-and-error to determine a single parameter for each augmentation, e.g., the probability of dropping 
nodes. LG2AR, on the other hand, learns the sampling distribution over $\mathcal{T}$ and also learns parametric augmentation $\tau_{\phi_i}$ end-to-end
along with the representations to generate robust representations. The architecture of LG2AR (Figure \ref{fig:arch}) achieves this using an 
augmentation encoder, a probabilistic policy, a set of probabilistic augmentation heads, and a shared base encoder. The details are as follows. 

\subsection{Augmentation Encoder}
Augmentation encoder $g_\omega(.): \mathbb{R}^{|\mathcal{V}| \times d_x} \times \mathbb{R}^{|\mathcal{E}|} \longmapsto \mathbb{R}^{|\mathcal{V}| \times d_h} \times \mathbb{R}^{d_h}$ learns a set of node encoding $\mathbf{H}_v \in \mathbb{R}^{|\mathcal{V}| \times d_h}$ and a graph encoding $h_g \in \mathbb{R}^{d_h}$ 
over the input graph $G_k$ to provide the subsequent modules with expressive encodings so that they can condition their predictions on the input 
graphs. The augmentation encoder consists of a GNN producing node representations, a read-out function (summation) aggregating the 
representations into a graph representation, and two dedicated projection heads (three layer MLPs) applied to the learned representations. To 
encode graphs, we opted for expressive power and adopted graph isomorphism network (GIN) \citep{xu_2019_iclr}.

\subsection{Policy}

Policy $r_{\mu}(.) : \mathbb{R}^{|\mathcal{B}| \times d_h } \longmapsto \mathbb{R}^{|\tau |}$ is a probabilistic module that receives a batch of graph-level representations $\textbf{H}_g \in \mathbb{R}^{|\mathcal{B}| \times d_h }$ from the augmentation
encoder, constructs a categorical distribution over the possible augmentations $\mathcal{T}$, and then samples two augmentations, $\tau_{\phi_i}$ and $\tau_{\phi_j}$ from that 
distribution for each batch with temperature $t$. It is shown that conditioning the augmentation sampling on the dataset helps achieve better 
performance \citep{you2020graph}. However, feeding the whole dataset to the policy module is computationally expensive and hence we approximate 
it by conditioning the policy on mini-batches. Moreover, the policy must be invariant to the order of representations within the batch. To enforce this, we 
tried two strategies: (1) a policy instantiated as a deep set \citep{NIPS2017_f22e4747} where representations are first projected and then aggregated into 
a \textit{batch representation}, and (2) a policy instantiated as an RNN where we impose an order on the representations by sorting them based on 
their $L_2$-norm and then feeding them into a GRU \citep{cho-2014-learning}. We use the last hidden state as the \textit{batch representation}. We observed that 
GRU policy performed better. The policy module automates the ad-hoc trial-and-error augmentation selection process. To let the gradients flow 
back to the policy module, we use a skip-connection and multiply the final graph representations by their augmentation probabilities 
predicted by the policy.

\subsection{Augmentations}
We use five graph augmentations including three topological augmentations: \emph{node dropping},\emph{edge perturbation}, and 
\emph{sub-graph inducing}, one feature augmentation: \emph{feature masking}, and one \emph{identity augmentation}. These 
augmentations enforce the priors that the semantic meaning of a graph should not change due to perturbations applied to its features or
topology. We do not use compute-intensive augmentations such as graph diffusion. Unlike previous works in which the 
parameters of the augmentations are chosen either randomly or by heuristics, we opt to learn them end-to-end. For example, rather than 
dropping nodes randomly\citep{you2020graph} or computing the probability proportional to a centrality measure \citep{zhu2021graph}, we train 
a model to predict the distribution over all nodes within a graph and then sample from it to decide which nodes to drop. Unlike the policy 
module, the augmentations are conditioned on the individual graphs. We use a dedicated head for each augmentation modeled as a two 
layer MLP that learns a distribution over the augmentation parameters. The inputs to the heads are the original graph $G$ and representations
$\mathbf{H}_v$ and $h_G$ from the augmentation encoder. We sample the learned distribution using Gumbel-Softmax trick 
\citep{jang2016categorical, maddison2016concrete} with temperature $t$.

\textbf{Node Dropping Head} (Algorithm \ref{algo:node}) is conditioned on the node and graph representations to decide which nodes within a graph to drop. It receives the 
concatenation of the node and graph representations as input and predicts a categorical distribution over the nodes. The distribution is 
then sampled using Gumbel-Top-K trick \citep{kool2019stochastic} with a ratio hyper-parameter. We also tried Bernoulli sampling but we 
observed that it aggressively drops nodes in the few first epochs and the model cannot recover later. To let the gradients flow back from 
the augmented graph to the head, we introduce edge weights on the augmented graph where an edge weight $w_{ij}$ is computed as $p(v_i) + p(v_j)$ 
and $p(v_i)$ is the probability assigned to node $v_i$.

\textbf{Edge Perturbation Head} (Algorithm \ref{algo:edge}) is conditioned on head and tail nodes to decide which edges to add/drop. To achieve this, $|\mathcal{E}|$ 
\emph{negative edges} ($\bar{\mathcal{E}}$) are first randomly sampled to form a set of negative and positive edges  $\mathcal{E} \cup \bar{\mathcal{E}}$ with size of $2|\mathcal{E}|$. Edges represented as 
$\left[h_{v_i} + h_{v_j} \parallel \mathbbm{1}_{\mathcal{E}}(e_{ij})\right]$($h_{v_i}$ and $h_{v_j}$are the representations of head and tail nodes of edge $e_{ij}$ and $\mathbbm{1}_{\mathcal{E}}(e_{ij})$ indicates if the edge belongs to positive or negative edges) are fed into the head to learn Bernoulli distributions over the edged. We use the predicted 
probabilities $p(e_{ij})$ as the edge weights to let the gradients flow back to the head. 

\textbf{Sub-graph Inducing Head} (Algorithm \ref{algo:subgraph}) is conditioned on the node and graph representations to decide the center node. It receives the 
concatenation of the node and graph representations (i.e., $\left[h_v \parallel h_g\right]$) as input and learns a categorical distribution over the nodes. The distribution 
is then sampled to select a central node per graph around which a sub-graph is induced using Breadth-First Search (BFS) with K hops. We use a similar 
trick to node dropping augmentation to overpass the gradients back to the original graph.

\textbf{Feature Masking Head} (Algorithm \ref{algo:feature}) is conditioned on the node representation to decide which dimensions of the node feature to mask. The head receives the
node representation $h_v$ and learns a Bernoulli distribution over each feature dimension of the original node feature. The distribution is then sampled 
to construct a binary mask $m$ over the initial feature space. Because initial node features can consist of categorical attributes, 
we use a linear layer to project them into a continuous space resulting in $x'_v$. The augmented graph has the same structure as the original graph with 
initial node features $x'_v \odot m$ ($\odot$ is the Hadamard product). 

\subsection{Base Encoder}
Base encoder $g_\theta(.): \mathbb{R}^{|\mathcal{V}'| \times d_x'} \times \mathbb{R}^{|\mathcal{V'}| \times |\mathcal{V'}|} \longmapsto \mathbb{R}^{|\mathcal{V'}| \times d_h} \times \mathbb{R}^{d_h}$ is a shared graph encoder among the augmentations which receives an augmented 
graph $G'=\left( \mathcal{V}', \mathcal{E}'\right)$ from the corresponding augmentation head and learns a set of node representations $\mathbf{H}'_v \in \mathbb{R}^{|\mathcal{V}'| \times d_h}$ and a graph representation 
$h'_G \in \mathbb{R}^{d_h}$ over the augmented graph $G'$. The goal of learning the augmentations is to help the base encoder learn invariances to such augmentations 
and as a result produce robust representations. The base encoder is trained end-to-end with the policy and augmentation heads. At inference time, 
the input graphs are directly fed to the base encoder to compute the encodings for down-stream tasks.

\subsection{Training}
We follow \citep{Sun2020InfoGraph} and train the framework end-to-end using deep InfoMax \citep{hjelm_2019_iclr} objective: 
\begin{equation}
\max_{\omega, \mu \phi_i, \phi_j, \theta} \frac{1}{|\mathcal{G}|} \sum_{G \in \mathcal{G}} \left[ \frac{1}{|\mathcal{V}|} \sum_{v \in \mathcal{V}} \left[ \text{I}\left( h_v^{i}, h_G^j \right) + \text{I}\left( h_v^j, h_G^i \right)\right] \right]
\end{equation}
where ${\omega, \mu, \phi_i, \phi_j, \theta}$ are parameters of modules to be learned, $h_v^i, h_G^j$ are representations of node $v$ and graph $G$ encoded by augmentation $i$ and $j$, and $I$ is
the mutual information estimator. We use the Jensen-Shannon MI estimator:
\begin{align*} 
\text{I}\left(h_v, h_G \right) = \mathbb{E}_p \left[-\text{log} \left( 1 + \text{exp} \left(-\mathcal{D}(h_v, h_G)\right) \right)\right] - \\
 \mathbb{E}_{p \times \tilde{p}} \left[ -\text{log} \left( 1 + \text{exp} \left(\mathcal{D}(h_v, h_{\tilde{G}})\right) \right)\right]
\end{align*}
$\mathcal{D}(., .):\mathbb{R}^{d_h} \times \mathbb{R}^{d_h} \longmapsto \mathbb{R}$ is a discriminator that takes in a node and a graph representation and scores the agreement between them and is implemented 
as $\mathcal{D}(h_v, h_g)=h_n.h_g^T$. We provide the positive samples from the joint distribution $p$ and the negative samples from the product of marginals $p \times \tilde{p}$, and  
optimize the model parameters with respect to the objective using mini-batch stochastic gradient descent. We found that regularizing the encoders by
randomly alternating between training the base and augmentation encoders helps the base encoder to generalize better. For this purpose, we train the 
policy and the augmentation heads at each step, but we sample from a Bernoulli to decide whether to update the weights of the base or augmentation 
encoders . The training process is summarized in Algorithm \ref{algo:train}.

\begin{algorithm}[tb]
   \caption{End-to-end training algorithm.}
   \label{algo:train}
\begin{algorithmic}
   \STATE {\bfseries Input:} Augmentation heads $\tau_i, \tau_j \in \mathcal{T}$, policy $r$, graph encoders $g_\theta$ and $g_\omega$, MI estimator $\mathcal{I}$, loss $\mathcal{L}$, and graphs $G \in \mathcal{G}$
   \FOR{sampled batch $\mathcal{B}=\{G_k\}_{k=1}^{N} \in \mathcal{G}$}
   \STATE $\{(\textbf{H}_{v_k}, h_{G_k})\}_{k=1}^{N} = g_\omega\left(G_k\right)$
   \hfill\scriptsize\COMMENT{Batch encodings}\normalsize
   \STATE $i, j, p_i, p_j =r_{\mu}(\{(\textbf{H}_{v_k}, h_{G_k})\}_{k=1}^{N})$
   \hfill\scriptsize\COMMENT{Sample policy}\normalsize
   \FOR{$k=1$ {\bfseries to} $N$}
   \STATE $ G_k^i=\tau_{\phi_i}(G_k, \textbf{H}_{v_k}, h_{G_k})$
   \hfill\scriptsize\COMMENT{Sample 1st view}\normalsize
   \STATE $\textbf{H}_{v_k}^i, h_{G_k}^i = g_\theta(G_k^i)$ 
   \hfill\scriptsize\COMMENT{1st view encoding}\normalsize
   \STATE $ G_k^j=\tau_{\phi_j}(G_k, \textbf{H}_{v_k}, h_{G_k})$
   \hfill\scriptsize\COMMENT{Sample 2nd view}\normalsize
   \STATE $\textbf{H}_{v_k}^j, h_{G_k}^j = g_\theta(G_k^j)$
   \hfill\scriptsize\COMMENT{2nd view encoding}\normalsize
   \STATE $h_{G_k}^i = h_{G_k}^i \times p_i$
   \hfill\scriptsize\COMMENT{Scale encodings}\normalsize
   \STATE $h_{G_k}^j = h_{G_k}^j \times p_j$
   \ENDFOR
   \hfill\scriptsize\COMMENT{Compute pairwise similarity}\normalsize
   \FOR{$k=1$ {\bfseries to} $N$ And $k'=1$ {\bfseries to} $N$}
   \STATE $s_{k,k'}^i = \mathcal{I}\left(h_{G_k}^i, \textbf{H}_{v_{k'}}^j\right)$, $s_{k,k'}^j = \mathcal{I}\left(h_{G_k}^j, \textbf{H}_{v_{k'}}^i\right)$
   \ENDFOR
   \hfill\scriptsize\COMMENT{Compute gradients and update}\normalsize
   \STATE $\bigtriangledown_{\omega, \mu, \phi_i, \phi_j, \theta} \frac{1}{N^2} \sum\limits_{k=1}^{N} \sum\limits_{k'=1}^N{ \left[\mathcal{L}\left(s_{k,k'}^i\right) + \mathcal{L}\left(s_{k,k'}^j\right)\right]} $
   \ENDFOR
\end{algorithmic}
\end{algorithm}

\section{Experimental Results}
\subsection{Unsupervised Representation Learning Evaluation}
We evaluate LG2AR under the linear evaluation protocol on both node-level and graph-level classification benchmarks where unsupervised 
models first encode the graphs, and then the encoding are fed into a down-stream linear classifier without fine-tuning the encoder. For
graph classification benchmarks, we follow GraphCL and use eight datasets from TUDataset \citep{Morris2020} and for the node classification,
we follow GCA, and use seven datasets from \citep{mernyei2020wiki, shchur2018pitfalls}. For fair comparisons, we closely follow the linear 
evaluation protocol from previous unsupervised works. For evaluating graph classification, we report the mean 10-fold cross validation accuracy 
with standard deviation after ten runs with a down-stream linear SVM classifier, and for node classification evaluation, we report the mean accuracy 
of twenty runs over different data splits with a down-stream single layer linear classifier. For details on the evaluation protocol see Appendix
\ref{appndx:evaldetail} and for implementation details and hyper-parameter settings see Appendix \ref{appndx:ImpDetails}.

For both tasks, we train three variants of our framework denoted as LG2AR-GRU, LG2AR-DeepSet, and LG2AR-Random where each variant
indicates its policy instantiation, i.e., GRU, deep set, and random (sampling views from uniform distribution) policies. For graph classification
benchmarks, we compare the LG2AR with \emph{two supervised baselines}: GIN and Graph Attention Network (GAT) \citep{velickovic_2018_iclr}, 
\emph{five graph kernel methods} including Shortest Path kernel (SP) \citep{borgwardt_2005_icdm}, Graphlet Kernel (GK) \citep{shervashidze_2009_ais}, 
Weisfeiler-Lehman sub-tree kernel (WL) \citep{shervashidze_2011_jmlr}, Deep Graph Kernels (DGK) \citep{yanardag_2015_kdd}, and Multi-scale Laplacian 
Graph kernel (MLG) \citep{kondor_2016_nips}, and \emph{four random walk methods} including Random Walk \citep{gartner_2003_ltkm}, Node2Vec 
\citep{grover_2016_kdd}, Sub2Vec \citep{adhikari_2018_pakdd}, Graph2Vec \citep{narayanan_2017_arxiv}. We also compare the results with state-of-the-art deep 
contrastive models including InfoGraph, GraphCL, JOAO, and ADGCL. For node classification benchmarks, we compare our results with random walk 
methods including DeepWalk with and without concatenating node features, and Node2Vec. We also compare the results with deep learning methods 
including Graph Autoencoders (GAE, VGAE) \citep{kipf_2016_arxiv}, Graphical Mutual Information Maximization (GMI) \citep{peng2020}, MVGRL, DGI, and 
GCA. 

The results for graph classification are reported in Tables \ref{table:graph} and \ref{table:mix} and for the node classification are reported in 
Tables \ref{table:node} and \ref{table:mix}. The reported performance for other models are from the corresponding papers under the same 
experiment setting. As shown for graph classification, LG2AR achieves state-of-the-art results with respect to unsupervised models across all eight 
benchmarks. For example, on IMDB-Multi and COLLAB datasets we achieve 2.2\% and 6.4\% absolute improvement over previous state-of-the-art. 
We also observe that LG2AR narrows the gap with the best performing supervised baselines even surpassing them on MUTAG dataset. We also achieve state-of-the-art results for node classification in five out of seven benchmarks. 
For example, we achieve 4.7\%, 1.6\%, and 1.7 \% absolute improvement on the PubMed, Amazon-Photo, and Amazon-Computer datasets. When 
compared to supervised baselines, we outperform or perform equally good across the benchmarks. 

\begin{table*}[h!]
\caption{Mean graph classification accuracy over 10 runs under linear evaluation protocol.}
\label{table:graph}
\begin{center}
\begin{small}
\begin{tabular}{clcccccc}
\toprule
\multicolumn{2}{c}{\textbf{Method}} & \textbf{Mutag} & \textbf{Proteins} & \textbf{IMDB-B} &\textbf{IMDB-M} & \textbf{Reddit-B} & \textbf{Reddit-M}\\
\midrule
\multirow{2}{*}{\begin{turn}{90}Sup.\end{turn}}
& GCN & 85.6$\pm$5.8 &  75.2$\pm$3.6   & 74.0$\pm$ 3.4 & 51.9$\pm$3.8 & 50.0$\pm$0.0 &  20.0$\pm$0.0          \\
& GIN & 89.4$\pm$5.6 &  76.2$\pm$2.8  & 75.1$\pm$5.1 &52.3$\pm$2.8 & 92.4$\pm$2.5 &  57.6$\pm$1.5        \\
& GAT & 89.4$\pm$6.1 &  74.7$\pm$4.0  & 70.5$\pm$2.3 & 47.8$\pm$3.1 & 85.2$\pm$3.3 &  45.9 $\pm0.1$  \\
\midrule
\multirow{5}{*}{\begin{turn}{90}Kernel\end{turn}}  
& SP        & 85.2$\pm$2.4 &  $-$          &   55.6$\pm$0.2 & 38.0$\pm$0.3 & 64.1$\pm$0.1 & 39.6$\pm$0.2\\
& GK        & 81.7$\pm$2.1 &  $-$          &   65.9$\pm$1.0 & 43.9$\pm$0.4 & 77.3$\pm$0.2 & 41.0$\pm$0.2\\
& WL        & 80.7$\pm$3.0 &  72.9$\pm$0.6 &  72.3$\pm$3.4 & 47.0$\pm$0.5 & 68.8$\pm$0.4 & 46.1$\pm$0.2\\
& DGK       & 87.4$\pm$2.7 &  73.3$\pm$0.8 &  67.0$\pm$0.6 & 44.6$\pm$0.5 & 78.0$\pm$0.4 & 41.3$\pm$0.2\\
& MLG       & 87.9$\pm$1.6 &  $-$          &   66.6$\pm$0.3 & 41.2$\pm$0.0 & $-$          & $-$ \\
\midrule
\multirow{4}{*}{\begin{turn}{90}Rnd Walk\end{turn}}
& RandomWalk        & 83.7$\pm$1.5  &  $-$          &  50.7$\pm$0.3 & 34.7$\pm$0.2 & $-$          & $-$\\
& Node2Vec  & 72.6$\pm$10.2 &  57.5$\pm$3.6 &  $-$          & $-$          & $-$          & $-$\\
& Sub2Vec   & 61.1$\pm$15.8 &  53.0$\pm$5.6 &  55.3$\pm$1.5 & 36.7$\pm$0.8 & 71.5$\pm$0.4 & 36.7$\pm$0.4 \\
& Graph2Vec & 83.2$\pm$9.6  &  73.3$\pm$2.1 &  71.1$\pm$0.5 & 50.4$\pm$0.9 & 75.8$\pm$1.0 & 47.9$\pm$0.3 \\
\midrule
\multirow{7}{*}{\begin{turn}{90}Unsupervised\end{turn}}
& InfoGraph & 89.0$\pm$1.1  &  74.4$\pm$0.3 &  73.0$\pm$0.9 & 49.7$\pm$0.5 & 82.5$\pm$1.4 & 53.5$\pm$1.0 \\ 
& GraphCL   & 86.8$\pm$1.4  &  74.4$\pm$0.5 &   71.1$\pm$0.4 & 48.5 ± 0.6        & 89.5$\pm$0.8 & 56.0$\pm$0.3 \\
& ADGCL & 89.7$\pm$1.0 & 73.8$\pm$0.5 &  71.6$\pm$1.0 & 49.9$\pm$0.7 & 85.5$\pm$0.8 & 54.9$\pm$0.4 \\
& JOAO      & 87.7$\pm$0.8  &  74.6$\pm$0.4 &  70.8$\pm$0.3 & $-$          & 86.4$\pm$1.5 & 56.0$\pm$0.3 \\
& LG2AR + GRU (Ours)      & \textbf{90.0$\pm$0.6}  & \textbf{75.0$\pm$0.5}  & \textbf{74.5$\pm$0.6}  & \textbf{51.9$\pm$0.3} & 91.8$\pm$0.4 &  \textbf{56.3$\pm$0.2}             \\
& LG2AR + DeepSet (Ours)      & 88.9$\pm$0.6  & 74.8$\pm$0.5  & 74.1$\pm$0.2  & 51.2$\pm$0.3 & 91.6$\pm$0.1 &  56.0$\pm$0.2             \\
& LG2AR + Random (Ours)      & 88.6$\pm$0.5  & 74.7$\pm$0.5  & 73.8$\pm$0.3  & 51.5$\pm$0.3 & \textbf{92.2$\pm$0.1} &  56.2$\pm$0.2             \\
\bottomrule
\end{tabular}
\end{small}
\end{center}
\end{table*}

\begin{table*}[h!]
\caption{Mean node classification accuracy over 20 runs under linear evaluation protocol.}
\label{table:node}
\begin{center}
\begin{small}
\begin{tabular}{clccccc}
\toprule
\multicolumn{2}{c}{\textbf{Method}} & \textbf{WikiCS} & \textbf{Amz-Comp} & \textbf{Amz-Photo} & \textbf{Coauthor-CS} &  \textbf{Coauthor-Phy}\\
\midrule
\multirow{2}{*}{\begin{turn}{90}Sup.\end{turn}}
& GCN &  77.2$\pm$0.1 & 86.5$\pm$0.5 & 92.4$\pm$0.2 & 93.0$\pm$0.3 & 95.7$\pm$0.2 \\
& GIN & 75.9$\pm$0.7 & 87.5$\pm$0.9 & 90.9$\pm$0.5 & 91.4$\pm$0.2 & 95.2$\pm$0.1 \\
& GAT &  77.7$\pm$0.1 & 86.9$\pm$0.3 & 92.6$\pm$0.4 & 92.3$\pm$0.2 & 95.5$\pm$0.2 \\
\midrule
\multirow{4}{*}{\begin{turn}{90}Rnd Walk\end{turn}}  
& Raw Features &  71.98$\pm$0.0 & 73.8$\pm$0.0 & 78.5$\pm$0.0 & 90.4$\pm$0.0 & 93.6$\pm$0.0 \\
& Node2Vec &  71.8$\pm$0.1 & 84.4$\pm$0.1 & 89.7$\pm$0.1 & 85.1$\pm$0.0 & 91.2$\pm$0.0 \\
& DeepWalk &  74.4$\pm$0.1 & 85.7$\pm$0.1 & 89.4$\pm$0.1 & 84.6$\pm$0.2 & 91.8$\pm$0.2 \\
& DeepWalk + Feat &  77.2$\pm$0.0 & 86.3$\pm$0.1 & 90.1$\pm$0.1 & 87.7$\pm$0.0 & 94.9$\pm$0.1 \\
\midrule
\multirow{11}{*}{\begin{turn}{90}Unsupervised\end{turn}}
& GAE    &  70.2$\pm$0.0 & 85.3$\pm$0.2 & 91.6$\pm$0.1 & 90.0$\pm$0.7 & 94.92$\pm$0.1 \\
& VGAE   &  75.6$\pm$0.2 & 86.4$\pm$0.2 & 92.2$\pm$0.1 & 92.1$\pm$0.1 & 94.5$\pm$0.0 \\
& DGI    &  75.4$\pm$0.1 & 84.0$\pm$0.5 & 91.6$\pm$0.2 & 92.2$\pm$0.6 & 94.5$\pm$0.5 \\
& GMI    &  74.9$\pm$0.1 & 82.2$\pm$0.3 & 90.7$\pm$0.2 & OOM & OOM \\
& MVGRL  &  77.5$\pm$0.0 & 87.5$\pm$0.1 & 91.7$\pm$0.1 & 92.1$\pm$0.1 & 95.3$\pm$0.0 \\
& GRACE   &  \textbf{80.1$\pm$0.5} & 89.5$\pm$0.4 & 92.8$\pm$0.5 & 91.1$\pm$0.2  & OOM \\
& BGRL  &  80.0$\pm$0.1 & \textbf{90.3$\pm$0.2} & 93.2$\pm$0.3 & 93.3$\pm$0.1 & \textbf{95.7$\pm$0.1} \\
& GCA       &  78.4$\pm$0.1 & 87.9$\pm$0.3 & 92.2$\pm$0.2 & 93.1$\pm$0.0 & \textbf{95.7$\pm$0.0} \\
& LG2AR + GRU (Ours)   &       77.8 $\pm$0.5 & 89.3$\pm$0.4 & \textbf{94.1$\pm$0.4} & \textbf{93.6$\pm$0.3} & \textbf{95.7$\pm$0.2} \\
& LG2AR + DeepSet (Ours)   & 76.2$\pm$0.9 & 89.6$\pm$0.3 & 92.6$\pm$0.5 & 92.4$\pm$0.3 & 95.5$\pm$0.1 \\
& LG2AR + Random (Ours)   &76.2$\pm$0.7 & 88.8$\pm$0.4 & 92.4$\pm$0.6 & 92.3$\pm$0.2 & 95.4$\pm$0.1 \\
\bottomrule
\end{tabular}
\end{small}
\end{center}
\end{table*}

\begin{table*}[h!]
\caption{Mean graph and node classification accuracy under linear evaluation protocol.}
\label{table:mix}
\begin{center}
\begin{small}
\begin{tabular}{lcc||lcc}
\toprule
\multicolumn{3}{c}{\textit{Node}} & \multicolumn{3}{c}{\textit{Graph}}\\
\midrule
\textbf{Method} &\textbf{Cora} &\textbf{PubMed} & \textbf{Method} &\textbf{Collab} &\textbf{DD}\\
\midrule
DeepWalk  & 70.7$\pm$0.6 &  74.3$\pm$0.9 & InfoGraph &  70.7$\pm$1.1 &  72.9$\pm$1.8  \\
GAE & 71.5$\pm$0.4 & 72.1$\pm$0.5 & GraphCL   &  71.4$\pm$1.2 &  78.6$\pm$0.4  \\
VERSE & 72.5$\pm$0.3  & $-$ & AD-GCL  &  73.3$\pm$0.6 &  75.1$\pm$0.4  \\
DGI & 82.3$\pm$0.6  & 76.8$\pm$0.6 & JOAO  &  69.5$\pm$0.4 &  77.4$\pm$1.2  \\
LG2AR + GRU (Ours) & \textbf{82.7$\pm$0.7} & 81.0$\pm$0.6 & LG2AR + GRU (Ours)      &\textbf{ 77.8$\pm$0.2}  & \textbf{79.1$\pm$0.3}  \\
LG2AR + DeepSet (Ours) & 80.8$\pm$1.0 & \textbf{81.5$\pm$0.7} & LG2AR + DeepSet (Ours)      & \textbf{77.8$\pm$0.2}  & 78.6$\pm$0.5  \\
LG2AR + Random (Ours)  & 81.6$\pm$0.9 & 81.3$\pm$0.8 & LG2AR + Random (Ours)       & 77.6$\pm$0.2  & 78.8$\pm$0.4  \\
\bottomrule
\end{tabular}
\end{small}
\end{center}
\end{table*}

\subsection{Semi-Supervised Learning Evaluation}
Furthermore, we evaluated LG2AR in a semi-supervised learning setting on graph classification benchmarks. We follow the experimental protocol introduced in GraphCL and pre-train the
encoder in an unsupervised fashion and fine-tune it only on 10\% of the labeled data. The results reported in Table \ref{table:semi} show that LG2AR achieves state-of-the-art
results compared to previous unsupervised models across all five benchmarks. Most notably, LG2AR achieves an absolute accuracy gain of 3.6\% and 2.6\% over Collab and Reddit-Multi 
benchmarks.

\begin{table*}
\caption{Mean 10-fold accuracy of semi-supervised learning with 10\% label rate.}
\label{table:semi}
\begin{center}
\begin{small}
\begin{tabular}{lccccc}
\toprule
& \textbf{Proteins} & \textbf{DD} & \textbf{COLLAB} & \textbf{ Reddit-B} & \textbf{Reddit-M} \\
\midrule
GAE           		  & 70.5$\pm$0.2 & 74.5$\pm$0.7 & 75.1$\pm$0.2 & 87.7$\pm$0.4 & 53.6$\pm$0.1 \\
Infomax     		& 72.3$\pm$0.4 & 75.8$\pm$0.3 & 73.8$\pm$0.3 & 88.7$\pm$0.9 & 53.6$\pm$0.3 \\
GraphCL   		   & 74.2$\pm$0.3 & 76.2$\pm$1.4 & 74.2$\pm$0.2 & 89.1$\pm$0.2 & 52.6$\pm$0.5 \\
JOAO                &  73.3$\pm$0.5 & 75.8$\pm$0.7 & 75.5$\pm$0.2 & 88.8$\pm$0.7 & 52.7$\pm$0.3 \\
ADGCL               &  74.0$\pm$0.5 & 77.9$\pm$0.7 & 75.8$\pm$0.3 & 90.1$\pm$0.2 & 53.5$\pm$0.3 \\
LG2AR (Ours)    & \textbf{76.1$\pm$0.4} &\textbf{ 79.8$\pm$0.3} & \textbf{78.4$\pm$0.4} & \textbf{92.3$\pm$0.5} & \textbf{57.2$\pm$0.6} \\
\bottomrule
\end{tabular}
\end{small}
\end{center}
\end{table*}

\subsection{Analysis of the optimization framework}
In this section, we discuss how LG2AR cannot fall into trivial solutions and compare its optimization with a few notable works. SimSiam \citep{chen2021exploring} states that collapsing, i.e. minimum possible loss with constant outputs, cannot be prevented by solely relying on architecture designs such as batch normalization. By designing multiple experiments they concluded that the non-collapsing behaviour of SimSiam still remains an empirical observation. Inspired by MoCo 
\citep{he2020momentum} and BYOL \citep{NEURIPS2020_f3ada80d}, AutoMix \citep{liu2021automix} avoids a nested optimization by decoupling its momentum pipeline. LG2AR does not require a momentum pipeline because it is based on contrasting local-global information. Even if the augmentation distribution ends being a Dirac peaking on any of the five augmentations, which we did not observe, a collapse cannot happen and a single level optimization is sufficient. If solely any of the node, sub-graph or edge augmentations are sampled, the inductive biases that we injected and discussed, force the model to at least select a sub part of the graph as an augmented view, hence avoiding the trivial solution of a graph with only one node and no edges. If only identity or feature augmentation is sampled, LG2AR would reduce to a single level optimization method such as InfoGraph where there are no augmentations.
JOAO, re-frames the auto-augmentation on graphs to a min-max optimization. Inspired from robustness and adversarial learning literature, they employ Alternating Gradient Descent \citep{wang2019towards} to design an approach for learning the augmentation distribution and the encoder parameters in a bi-level optimization setting. We are using the gumbel-softmax trick to sample from the augmentation distribution and let the gradient flow through discrete parameters. Our algorithm alternates between updating the encoder and augmentation parameters by tossing a fair coin in each iteration. Alternating gradients between modules in unsupervised learning is shown to be efficient in avoiding the trivial solutions 
\citep{caron2018deep, hassani2019unsupervised, Khasahmadi2020Memory-Based}. With these bag of tricks, LG2AR avoid collapsing to trivial solutions and solving a min-max problem that needs an inefficient bi-level optimization. Moreover,  Figure \ref{fig:loss} in the Appendix is showing a stable training trajectory for multiple datasets.

\begin{table*}
\caption{Effects of Mutual Information Estimator, Discriminator, and augmentations.}
\label{table:ablation}
\setlength{\tabcolsep}{4pt}
\begin{center}
\begin{small}
\begin{tabular}{llccccccc}
\toprule
& & \textbf{Proteins} & \textbf{DD} & \textbf{COLLAB} & \textbf{IMDB-B} & \textbf{IMDB-M} & \textbf{ Reddit-B} & \textbf{Reddit-M} \\
\midrule
\multirow{4}{*}{\begin{turn}{90}Loss\end{turn}}
& \textbf{\underline{JSD}} & 75.0$\pm$0.5 & \textbf{79.1$\pm$0.3} & \textbf{77.8$\pm$0.2} & 74.5$\pm$0.6 & \textbf{51.9$\pm$0.3} & \textbf{91.8$\pm$0.4} & 56.3$\pm$0.2 \\
& \textbf{NCE}             & 74.4$\pm$0.6 & 78.4$\pm$0.5 & 77.1$\pm$0.3 & 73.9$\pm$0.4 & 51.2$\pm$0.6 & 90.8$\pm$0.4 & 56.4$\pm$0.2 \\
& \textbf{NT-Xent}         & \textbf{75.1$\pm$0.6} & 78.7$\pm$0.5 & 77.5$\pm$0.4 & \textbf{74.6$\pm$0.7} & 51.5$\pm$0.4 & 91.3$\pm$0.5 & \textbf{56.7$\pm$0.4} \\
& \textbf{DV}              & 74.3$\pm$0.4 & 78.2$\pm$0.4 & 77.1$\pm$0.3 & 73.5$\pm$0.6 & 50.7$\pm$0.5 & 91.1$\pm$0.5 & 55.2$\pm$0.3 \\
\midrule
\multirow{4}{*}{\begin{turn}{90}Discrimi.\end{turn}}
& \textbf{\underline{Dot Product}} & 75.0$\pm$0.5 & 79.1$\pm$0.3 & \textbf{77.8$\pm$0.2} & 74.5$\pm$0.6  & 51.9$\pm$0.3 & \textbf{91.8$\pm$0.4} & 56.3$\pm$0.2 \\
& \textbf{Cosine}                  & 75.4$\pm$0.4 & 79.2$\pm$0.3 & 77.4$\pm$0.3 & 74.3$\pm$0.7  & 51.6$\pm$0.4 & 92.1$\pm$0.3 & 56.4$\pm$0.2 \\
& \textbf{Bilinear}                & 74.6$\pm$0.4 & 78.7$\pm$0.4 & 77.5$\pm$0.4 & 73.8$\pm$0.6  & 51.0$\pm$0.5 & 90.4$\pm$0.5 & 55.4$\pm$0.4 \\
& \textbf{MLP}                     & \textbf{75.3$\pm$0.6} & \textbf{79.6$\pm$0.5} & 77.5$\pm$0.5 & \textbf{74.7$\pm$0.3}  & \textbf{60.4$\pm$0.6} & 91.7$\pm$0.5 & \textbf{56.8$\pm$0.5} \\
\midrule
\multirow{6}{*}{\begin{turn}{90}Augmentations\end{turn}}
& \textbf{\underline{All}}  & \textbf{75.0$\pm$0.5} & \textbf{79.1$\pm$0.3} & \textbf{77.8$\pm$0.2} & \textbf{74.5$\pm$0.6}  & \textbf{51.9$\pm$0.3} & \textbf{91.8$\pm$0.4} & \textbf{56.3$\pm$0.2} \\
& \textbf{Structure}        & 74.6$\pm$0.5 & 79.1$\pm$0.2 & 77.3$\pm$0.3 & 74.1$\pm$0.5  & 51.8$\pm$0.3 & 91.1$\pm$0.3 & 56.2$\pm$0.2 \\
& \textbf{Feature}          & 74.2$\pm$0.4 & 77.9$\pm$0.3 & 76.7$\pm$0.3 & 73.7$\pm$0.2  & 51.3$\pm$0.3 & 89.1$\pm$0.4 & 55.3$\pm$0.3 \\
& \textbf{Node}             & 74.1$\pm$0.5 & 78.1$\pm$0.2 & 76.6$\pm$0.2 & 73.6$\pm$0.3  & 51.2$\pm$0.3 & 89.4$\pm$0.3 & 55.1$\pm$0.2 \\
& \textbf{Edge}             & 74.3$\pm$0.5 & 77.7$\pm$0.2 & 76.9$\pm$0.2 & 73.7$\pm$0.3  & 50.9$\pm$0.3 & 89.5$\pm$0.3 & 55.2$\pm$0.2 \\
& \textbf{SubGraph}         & 73.9$\pm$0.5 & 78.4$\pm$0.3 & 76.4$\pm$0.2 & 73.1$\pm$0.3  & 51.2$\pm$0.4 & 89.4$\pm$0.2 & 55.3$\pm$0.3 \\
\bottomrule
\end{tabular}
\end{small}
\end{center}
\end{table*}

\begin{figure*}[h!] 
\begin{center}
\centerline{\includegraphics[width=120mm]{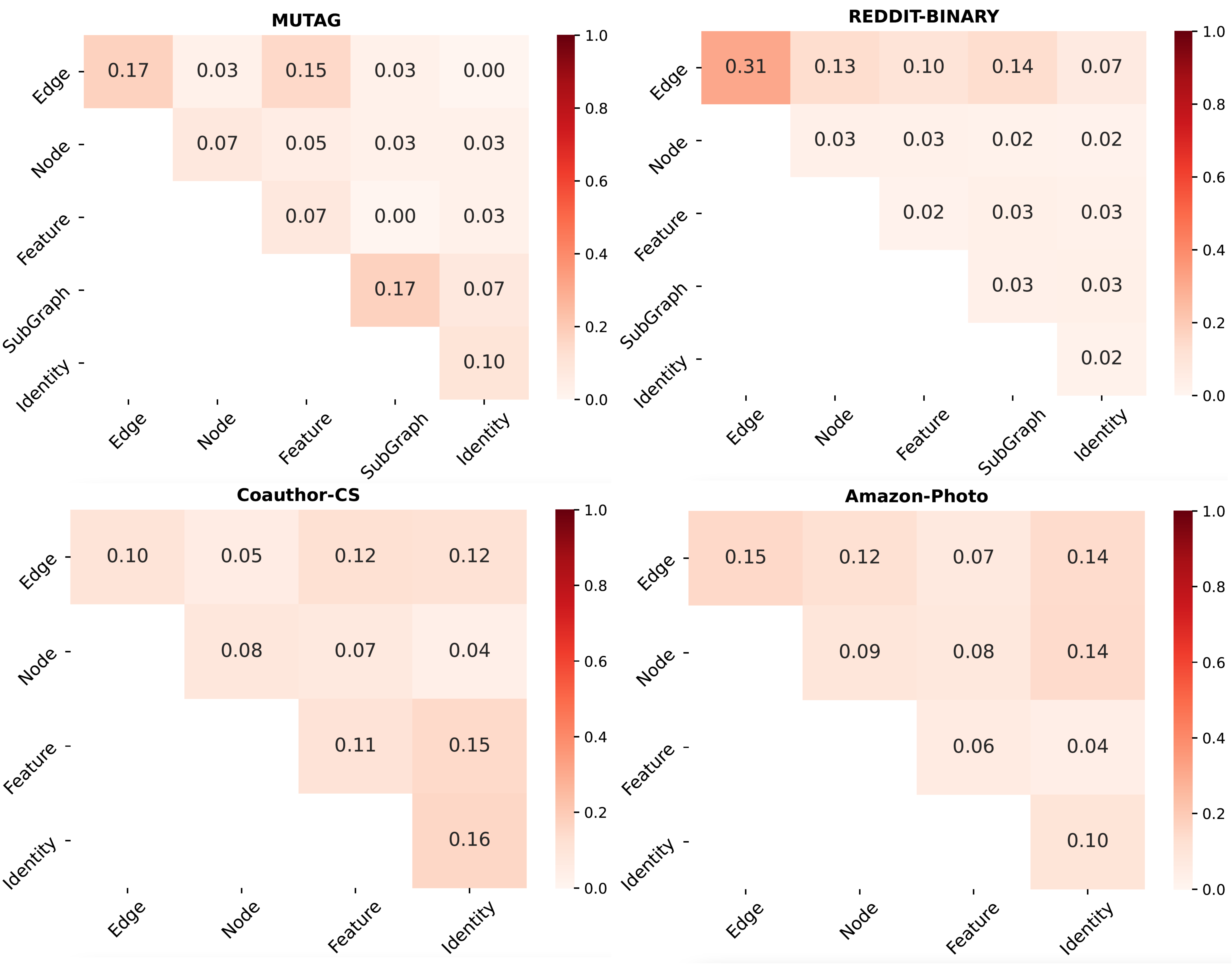}}
\caption{The normalized frequency of augmentation selection by the GRU-based policy for two graph benchmarks (top row) and two node 
benchmarks (bottom row).}\label{fig:confusion}
\end{center}
\end{figure*}

\subsection{Ablation Study }
\textbf{Effect of the Policy.} To investigate the effect of policy, we trained the models with three variants of the policy including GRU, deep set, and 
random policies. As shown in Tables \ref{table:graph}-\ref{table:node}, the GRU-based policy outperforms or performs equally well on 12 out 
of 15 benchmarks whereas the random policy outperforms the other variants in only 1 out of 15 datasets, indicating the importance of learning 
the view sampling policy. Also, in order to probe what the policy module is learning, we computed the normalized frequency of the sampled 
augmentations by the GRU-based policy during the training. The frequencies for two graph classification benchmarks (MUTAG and 
Reddit-binary) and two node classification benchmarks (Coauthor-CS and PubMed) are shown in Figure \ref{fig:confusion}. We observe the 
following: (1) The policy is learning different distributions over augmentations for each benchmark suggesting that it is adapting to the given 
datasets. (2) In node classification benchmarks, because we already induce sub-graphs to transform them to inductive tasks, we observe that 
the policy inclines towards sampling the identity augmentation more frequently which is essentially a sub-graph of the original graph. (3) We 
observe that regardless of the task, edge perturbation and sub-graph inducing are the two most commonly sampled augmentations. This 
confirms the observation that sub-graphs are generally beneficial across datasets \citep{you2020graph}. (4) We observe that for node classification 
benchmarks, the probability of sampling feature masking augmentation is positively correlated with the initial node feature dimension.

\textbf{Effect of the Augmentations.} To investigate the effect of the augmentations, we run the experiments with single augmentation, and structural vs 
feature space augmentations. The results shown in Table \ref{table:ablation} indicate that: (1) using our single augmentations performs on par or better than baselines, (2) generally structural augmentations contribute more than feature space augmentations, and (3) all augmentations are contributing to the final performance which suggests that the model should use all augmentations while learning the sampling frequencies.

\textbf{Effect of the Augmentation Heads.} To investigate the effect of the augmentations heads without benefiting from the policy, we compare our 
framework when trained with a random policy with GraphCL for graph classification benchmarks. Both LG2AR-Random and GraphCL sample 
the augmentations from a uniform distribution, where the former learns distributions over the augmentation parameters and the latter 
randomly samples those. As shown in Tables \ref{table:graph} and \ref{table:mix}, in 8 out of 8 benchmarks LG2AR-Random outperforms GraphCL. 
For instance, we see an absolute improvement of 2.7\% accuracy on Reddit-Binary. This implies that learning the augmentations contributes to the 
performance boost on the graph classification benchmarks, and when combined with the policy learning, results are further improved. We see less 
of this effect in transductive tasks suggesting that the policy learning is playing a more important rule in node classification benchmarks. One 
reason for this may be the fact that GCA unlike GraphCL uses strong topological inductive biases to select the augmentation parameters. 

\textbf{Effect of the Mutual Information Estimator and Discriminator.} We investigated four MI estimators including: noise-contrastive estimation (NCE) 
\cite{gutmann_2010_aistat, oord_2018_arxiv}, Jensen-Shannon (JSD) estimator following formulation in \cite{nowozin_2016_nips}, normalized temperature-scaled 
cross-entropy (NT-Xent) \cite{chen_2020_arxiv}, and Donsker-Varadhan (DV) representation of the KL-divergence \cite{donsker_1975_comm}. The results shown in 
Table \ref{table:ablation} suggests that JSD and NT-Xent perform better compared to the other estimators. We also investigated the effect of discriminator 
by training the model using four variants including dot product, cosine distance, Bilinear, and MLP discriminators. The results shown in 
Table \ref{table:ablation} suggests that discriminator instantiated as an MLP performs better across the datasets.



\section{Conclusion}
We introduced LG2AR, and end-to-end framework to automate graph contrastive learning. The proposed framework learns the augmentations, 
view selection policy, and the encoders end-to-end without requiring ad-hoc trial-and-error processes for devising the augmentations for each and 
every dataset. Experimental results showed that LG2AR achieves state-of-the-art results on 8 out of 8 graph classification benchmarks, and 6 out of 
7 node classification benchmarks compared to the previous unsupervised methods. The results also suggest that LG2AR narrows the gap with its 
supervised counterparts. Furthermore, the results suggest that both learning the policy and learning the augmentations contributes to the performance.  
In future work, we are planning to investigate large pre-training and transfer learning capabilities of the proposed method.

\bibliography{icml2022}
\bibliographystyle{icml2022}

\newpage
\appendix
\onecolumn
\section{Appendix}

\subsection{Benchmarks} \label{appx:benchmarks}
We use seven node classification and eight graph classification benchmarks reported by previous state-of-the-art methods. 
For node classification benchmarks, we follow GCA \citep{zhu2021graph} and use Wiki-CS \citep{mernyei2020wiki} which is a computer 
science subset of the Wikipedia, Amazon-Computers and Amazon-Photo \citep{shchur2018pitfalls} which are networks of co-purchase 
relationships constructed from Amazon, Coauthor-CS and Coauthor-Physics \citep{shchur2018pitfalls} which are two academic networks 
containing co-authorship graphs, and two other citation networks, Cora and Pubmed \cite{sen_2008_aimag}. For graph classification 
benchmarks, we follow GraphCL \citep{you2020graph} and use benchmarks from TUDatasets \citep{Morris2020}. We use Proteins and DD 
\citep{dobson2003distinguishing} modeling neighborhoods in the amino-acid sequences and protein structures, respectively, MUTAG 
\cite{kriege_2012_icml} modeling compounds tested for carcinogenicity, COLLAB \citep{yanardag_2015_kdd} derived from 3 public 
physics collaboration, Reddit-Binary and Reddit-Multi-5K \citep{yanardag_2015_kdd} connecting users through responses in Reddit 
online discussions, and IMDB-Binary and IMDB-Multi \cite{yanardag_2015_kdd} connecting actors/actresses  based on movie appearances.
The statistics of the graph and graph classification benchmarks are summarized in Tables \ref{table:graphstat} and \ref{table:nodestat}, 
respectively.

\begin{table}[h!]
\caption{Statistics of graph classification benchmarks.}
\label{table:graphstat}
\setlength{\tabcolsep}{3.pt}
\begin{center}
\begin{small}
\begin{sc}
\begin{tabular}{lccc|ccccc}
\toprule
& \multicolumn{3}{c}{\textit{Biology}} & \multicolumn{5}{c}{\textit{Social Networks}}\\
\cline{2-9}
&                  \textbf{mutag} & \textbf{proteins} &  \textbf{dd}&  \textbf{collab} & \textbf{imdb-b} & \textbf{imdb-m} &\textbf{ reddit-b} & \textbf{reddit-m}\\
\midrule
$|$ Graphs$|$   & 188	     & 1113	    & 1178     &  5000 		& 1000	 & 1500	  &	2000       & 4999		\\			
$|$ Nodes$|$    & 17.93   &  39.06  &   284.32  &  74.49	  &  19.77  & 13.00  & 429.63   &  508.52   \\
$|$ Edges$|$     & 19.79   &  72.82  &   715.66  &  2457.78	 &  96.53  & 65.94  & 497.75   & 594.87    \\
$|$ Features$|$ & 7  &  4  &   89  &  0	 &  0 & 0  & 0   & 0    \\
$|$ Classes$|$   & 2          & 2	         &   2            &  3			   & 2         & 3          & 2	          & 5               \\
\bottomrule
\end{tabular}
\end{sc}	
\end{small}
\end{center}
\end{table}

\begin{table}[h!]
\caption{Statistics of node classification benchmarks.}
\label{table:nodestat}
 \setlength{\tabcolsep}{3pt}
\begin{center}
\begin{small}
\begin{sc}
\begin{tabular}{lcccccccc}
\toprule
& \textbf{cora} & \textbf{pubmed} & \textbf{wikics} & \textbf{amz-comp} & \textbf{amz-photo} & \textbf{coau-cs} &  \textbf{coau-phy}\\
\midrule
$|$ Nodes$|$  & 3,327 &  19,717 & 11,701 & 13,752 & 7,650 & 18,333 & 34,493 \\
$|$ Edges$|$  & 4,732 &  44,338 & 216,123 &  245,861 &  119,081 &  81,894 & 247,962\\
$|$ Features$|$  & 1,433 & 500  & 300 &  767 &  745 &  6,805 & 8,415\\
$|$ Classes$|$  & 6 &  3 & 10 & 10 & 8 & 15 & 5 \\
\bottomrule
\end{tabular}
\end{sc}	
\end{small}
\end{center}
\end{table}

\subsection{Evaluation Protocol Details}\label{appndx:evaldetail}
For node classification, we follow \citep{velickovic_2019_iclr, zhu2021graph} where we train the model with the 
contrastive method, and then use the resulting embeddings to train and test a simple logistic regression classifier. 
We train the model for twenty runs over different data splits and report the mean accuracy with standard deviation. For fair evaluation and 
following GCA, we use a two-layer Graph Convolution Network (GCN) \citep{kipf_2017_iclr} for the base encoder across all node classification 
benchmarks. In order to make the transductive node classification benchmarks compatible with our inductive framework, we sample $|\mathcal{B}|$ 
sub-graphs from the input graph around randomly selected nodes to emulate a batch of graphs, and then feed the batch to our framework. 
Because sub-graph augmentation occurs before our framework, we remove this augmentation from the policy and also remove its 
corresponding head. For graph classification, we follow \citep{you2020graph,Sun2020InfoGraph} where we first use the contrastive loss to train 
the model and then report the best mean 10-fold cross validation accuracy with standard deviation after five runs. The classifier is a linear SVM  
trained using cross-validation on the training folds of the learned embeddings. Following \citep{Sun2020InfoGraph}, we use GIN layers for the 
base encoder and treat the number of layers as a hyper-parameter. We observed that contrasting graph-level representation achieves better 
results in graph classification benchmarks. Finally, we report more node and graph level evaluation results under linear evaluation protocol in 
Table \ref{table:mix}.

\subsection{Implementation \& Hyper-Parameter Selection}\label{appndx:ImpDetails}
We implemented the experiments using PyTorch and used Pytorch Geometric library to implement the graph encoders. Each experiments was run on 
a single RTX 6000 GPU. We initialize the parameters using Xavier initialization and train the model using Adam optimizer. All our graph implementations 
are sparse and in Pytorch Geometric format. Therefore, in order to let the gradients back-propagate, we use edge weights computed from augmentation heads 
as a way to pass the gradients. Also, in order to let the gradients back-propagate to the policy module, we multiply the final graph encodings from the two views 
with the associated probability of each view computed by the policy.

For node graph classification benchmarks, following GCA, we fix the base encoder to a tow-layer GCN model with mean-pooling as the read-out function. We select
the number of augmentation encoder layers from [1, 2, 3, 4, 5, 6], number of sub-graphs per batch from [4, 8, 12, 16, 32], hidden dimension from [128, 256, 512], learning
rate from [1e-4, 1e-1], number of hops from [1, 2, 3, 4, 5, 6], temperature from [0.7, 1.4], node dropping ration from [0.6, 0.9], and the dropout from [0.0, 0.2]. The 
augmentation consists of GIN layers with three layer projection heads and a summation read-out function. Following DGI, we use a early-stopping with patience of
50 steps. Following GCA, we train the linear model for 300 epochs with the learning rate of 1e-2.

For graph classification benchmarks, following InfoGraph, we design the both base and augmentation encoders with GIN layers, dedicated three-layer projection heads
for node and graph encodings, and a summation read-out function. We share the learning rate and the number of layers between the two encoders and select them 
from [1e-4, 1e-1] and [1, 2, 3, 4, 5, 6], respectively. We select the batch size from [32, 64, 128], hidden dimension from [128, 256, 512], number of epochs from 
[10, 20, 40, 60, 100, 200], learning rate from [1e-4, 1e-1], number of hops from [1, 2, 3, 4, 5, 6], temperature from [0.7, 1.4], node dropping ration from [0.6, 0.9], and the dropout 
from [0.0, 0.2]. We also follow InfoGraph for graph classification and choose the C parameter of the SVM from $ [10^{-3}$, $10^{-2}$, ..., $10^2$, $10^3$]. The selected hyper-parameters 
are shown in Table \ref{table:hyper}.

\begin{table}[h!]
\caption{Selected hyper-parameters.}
\label{table:hyper}
\setlength{\tabcolsep}{2.5pt}
\begin{center}
\begin{small}
\begin{tabular}{llccccccccc}
\toprule
& \textbf{Benchmark} &\textbf{$|$Hidden$|$} &\textbf{$|$Batch$|$} & \textbf{Epoch} & \textbf{$|$Layers$|$} &\textbf{Learning Rate} &\textbf{Temperature} & \textbf{$|$Hops$|$} & \textbf{Ratio} & \textbf{Dropout} \\
\midrule
\multirow{8}{*}{\begin{turn}{90}Graph Benchmarks\end{turn}}
& \textbf{MUTAG} & 256 & 128 & 20 & 6 & 0.001 & 1.27 & 5 & 0.75 & 0.10 \\
& \textbf{Proteins} & 256 & 64 & 40 & 3 & 0.0003 & 0.73 & 4 & 0.77 & 0.05 \\
& \textbf{DD} & 256 & 128 & 100 & 3 & 0.0008 & 1.05 & 1 & 0.77 & 0.15 \\
& \textbf{COLLAB} & 128 & 128 & 10 & 4 & 0.0003 & 1.05 & 2 & 0.73 & 0.05 \\
& \textbf{IMDB-B} & 256 & 128 & 200 & 2 & 0.0003 & 1.04 & 4 & 0.84 & 0.00 \\
& \textbf{IMDB-M} & 512 & 128 & 100 & 3 & 0.0002 & 1.00 & 3 & 0.83 & 0.20 \\
& \textbf{ Reddit-B} &  128 & 64 & 20 & 6 & 0.001 & 1.17 & 4 & 0.89 & 0.05 \\
& \textbf{Reddit-M} & 128 & 128 & 10 & 6 & 0.0004 & 1.24 & 2 & 0.72  & 0.05 \\
\midrule
\multirow{7}{*}{\begin{turn}{90}Node Benchmarks\end{turn}}
& \textbf{CORA} & 512 & 12 & NA & 2 &  0.03 & 1.19 &  6 & 0.86 & 0.20 \\
& \textbf{PubMed} & 512 & 16 &NA & 3 &  0.001 &  1.37 &  4 &  0.85 &  0.00 \\
& \textbf{WikiCS} & 512 & 8 &NA & 2 & 0.0001 & 0.83 & 1 & 0.86 & 0.05 \\
& \textbf{Amz-Comp} & 512 & 16 &NA & 2 & 0.0001 & 1.17 & 1 & 0.84  & 0.05 \\
& \textbf{Amz-Photo} & 512 & 16 & NA & 3 & 0.0005 & 1.23 & 1 & 0.76 & 0.20 \\
& \textbf{Coau-CS} &  256 & 4 & NA & 2 & 0.003 & 1.38 & 2 & 0.78 & 0.10 \\
& \textbf{Coau-Phy} & 512 & 8 & NA & 5 & 0.001 & 1.07 & 3 & 0.80 & 0.05 \\
\bottomrule
\end{tabular}
\end{small}
\end{center}
\end{table}

\begin{table}
\caption{Mean time (seconds per epoch) and space (Gigabytes of GPU memory) for Infograph (single encoder) and  LG2AR.}
\label{table:hyper}
\setlength{\tabcolsep}{2.5pt}
\vskip -0.1in
\begin{center}
\begin{small}
\begin{tabular}{llcccccccc}
\toprule
& &  \textbf{MUTAG} & \textbf{Proteins} & \textbf{DD} & \textbf{COLLAB} & \textbf{IMDB-B} & \textbf{IMDB-M} & \textbf{ Reddit-B} & \textbf{Reddit-M} \\
\midrule
\multirow{2}{*}{\begin{turn}{90}Info\end{turn}}
& \textbf{Time} (Sec/epoch) & 0.15 & 0.82 & 1.05 & 2.55 & 0.37 & 0.75 & 6.06 & 11.51 \\
& \textbf{Space} (Gigabytes) & 1.249 & 1.343  & 4.423 & 3.153 & 1.355 & 1.545 & 4.341 & 10.321 \\
\midrule
\multirow{2}{*}{\begin{turn}{90}Ours\end{turn}}
& \textbf{Time} (Sec/epoch)& 0.59 & 2.06 & 3.81 & 12.64 & 1.67 & 1.74 & 13.38 & 24.33 \\
& \textbf{Space} (Gigabytes) & 1.525 & 2.113 & 14.051 & 16.331 & 2.535 & 2.905 & 9.495 & 22.229 \\
\bottomrule
\end{tabular}
\end{small}
\end{center}
\end{table}

\newpage

\begin{algorithm}[tb]
   \caption{Sub-graph inducing head.}
   \label{algo:subgraph}
\begin{algorithmic}
   \STATE {\bfseries Input:} Node and graph encodings $\textbf{H}_v$ and $h_g$, graph $G=(\mathcal{V}, \mathcal{E} , \textbf{X})$, Number of hops $K$
   $p(\mathcal{V}) = \text{MLP}\left(\left[\textbf{H}_v \parallel h_g\right]\right) $
    \STATE $v_{center} \gets$ Sample-Categorical($p(\mathcal{V})$)
    \STATE $\mathcal{V'}, \mathcal{E'}, \mathbf{X'} \gets$ k-Hop-BFS($v_{center}, K$)
    \STATE $\mathcal{E'} \gets \mathcal{E} \subseteq \mathcal{V'} \times \mathcal{V'}$
    \STATE $\mathbf{W}_{\mathcal{E}} \gets \left[p(v_i) + p(v_j) \quad \forall e_{ij} \in \mathcal{E'}\right]$
    \STATE $G' \gets (\mathcal{V'}, \mathcal{E'}, \mathbf{X}', \mathbf{W}_{\mathcal{E}}) $
    \STATE {\bfseries Return }$G'$
\end{algorithmic}
\end{algorithm}

\begin{algorithm}[tb]
   \caption{Node dropping head.}
   \label{algo:node}
\begin{algorithmic}
   \STATE {\bfseries Input:} Node and graph encodings $\textbf{H}_v$ and $h_g$, graph $G=(\mathcal{V}, \mathcal{E} , \textbf{X})$, Ratio $\mu$
   \STATE    $p(\mathcal{V}) = \text{MLP}\left(\left[\textbf{H}_v \parallel h_g\right]\right) $
   \STATE $\mathcal{V'} \gets$ Sample-Top-K($p(\mathcal{V}), \mu$) 
   \STATE $\mathcal{E'} \gets \mathcal{E} \subseteq \mathcal{V'} \times \mathcal{V'}$ 
   \STATE $\mathbf{X'} \gets \mathbf{X}[\mathcal{V'}]$ 
   \STATE $\mathbf{W}_{\mathcal{E}} \gets \left[p(v_i) + p(v_j) \quad \forall e_{ij} \in \mathcal{E'}\right]$
   \STATE $G' \gets (\mathcal{V'}, \mathcal{E'}, \mathbf{X}', \mathbf{W}_{\mathcal{E}})$
   \STATE {\bfseries Return }$G'$
\end{algorithmic}
\end{algorithm}

\begin{algorithm}[tb]
   \caption{Edge perturbation head.}
   \label{algo:edge}
\begin{algorithmic}
   \STATE {\bfseries Input:} Node and graph encodings $\textbf{H}_v$ and $h_g$, graph $G=(\mathcal{V}, \mathcal{E} , \textbf{X})$, temperature $t$
   \STATE $\mathcal{V'}, \mathcal{E'}, \mathbf{X'}, \mathbf{W}_{\mathcal{E}}\gets \emptyset$
   \STATE $\mathcal{\bar{E}}=$ Sample-Negative-Edges($\mathcal{E}$)
    \FOR{$e_{ij}$ {\bfseries to} $\mathcal{\bar{E}} \cup \mathcal{E}$}
   \STATE $h_{e_{ij}} = \left[h_{v_i} + h_{v_j} \parallel \mathbbm{1}_{\mathcal{E}}(e_{ij})\right]$
   \STATE $p({e_{ij}}) = \text{MLP}(h_{e_{ij}})$
   \IF{Bernoulli-Sample($t, p({e_{ij}})$)}
   \STATE $\mathcal{V'} \gets \mathcal{V'} \cup \{v_i, v_j\}$
   \STATE $\mathcal{E'} \gets \mathcal{E'} \cup \{e_{ij}\}$
   \STATE $\mathbf{X'} \gets \mathbf{X'} \cup \{h_{v_i}, h_{v_j}\}$
   \STATE $\mathbf{W}_{\mathcal{E}} \gets \mathbf{W}_{\mathcal{E}} \cup \{p({e_{ij}}) \}$
   \ENDIF
   \STATE $G' \gets (\mathcal{V'}, \mathcal{E'}, \mathbf{X}', \mathbf{W}_{\mathcal{E}})$
  \ENDFOR
  \STATE {\bfseries Return }$G'$
\end{algorithmic}
\end{algorithm}

\begin{algorithm}[tb]
   \caption{Feature masking head.}
   \label{algo:feature}
\begin{algorithmic}
   \STATE {\bfseries Input:} Node and graph encodings $\textbf{H}_v$ and $h_g$, graph $G=(\mathcal{V}, \mathcal{E} , \textbf{X})$, temperature $t$
   \STATE $\textbf{X}' \gets \text{Linear}(\textbf{X})$
   \STATE $\textbf{M} \gets \text{Bernoulli-Sample(} \text{MLP}(\textbf{H}_v), t)$
   \STATE $\textbf{X}' \gets \textbf{X}' \odot \textbf{M} $
   \STATE $G' \gets (\mathcal{V}, \mathcal{E}, \mathbf{X}', \mathbf{W}_{\mathcal{E}}=\mathbf{1}) $
  \STATE {\bfseries Return }$G'$
\end{algorithmic}
\end{algorithm}

\begin{figure}
\begin{center}
\centerline{\includegraphics[width=150mm]{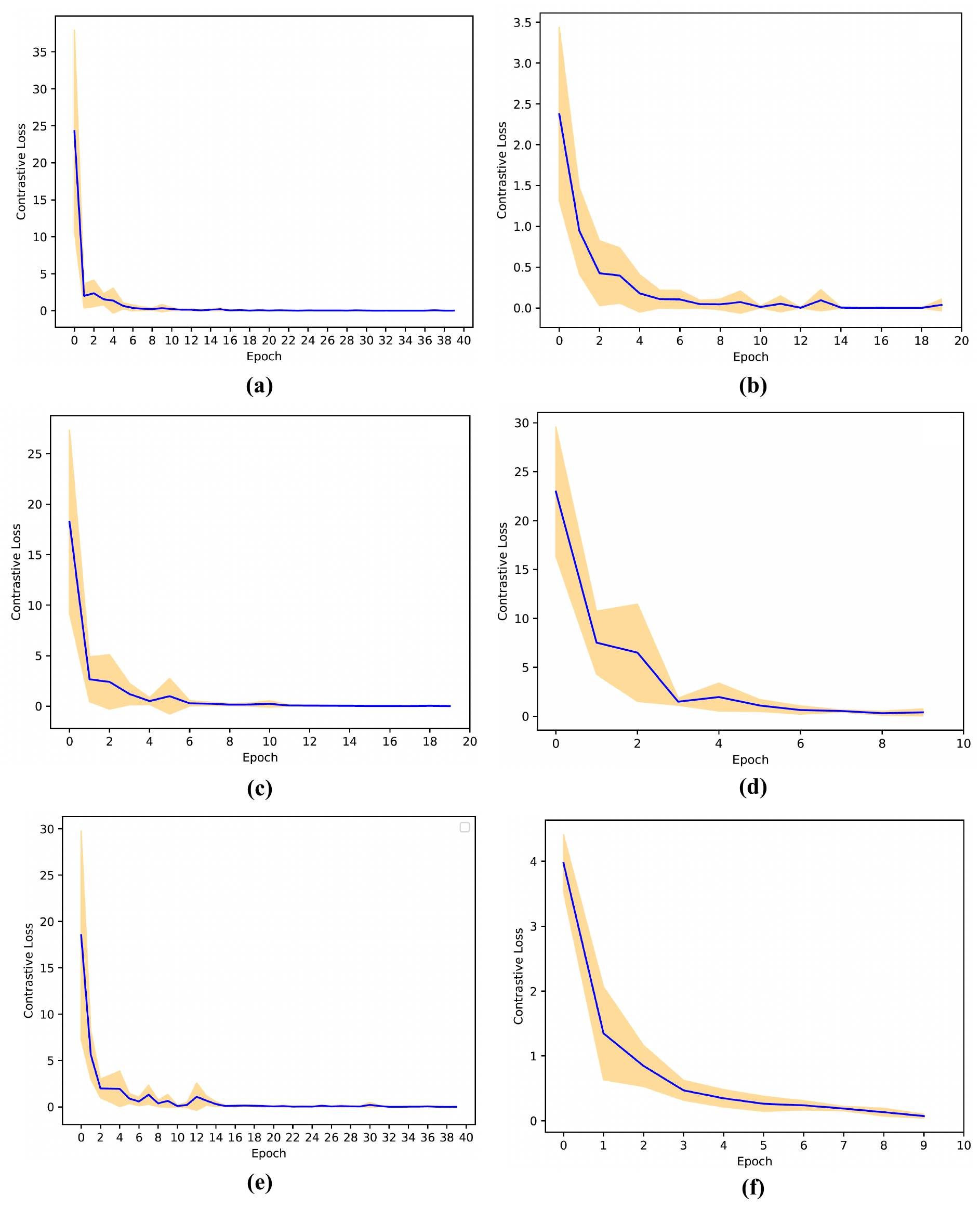}}
\caption{The evolution of the contrastive loss during the training averaged over ten runs for: (a) Proteins, (b) IMDB-Binary, 
(c) Reddit-Binary, (d) Reddit-Multi, (e) DD, and (f) Collab benchmarks.}\label{fig:loss}
\end{center}
\end{figure}

\end{document}